\documentclass[10pt, a4paper]{article}
\usepackage{lrec}
\usepackage{multibib}
\newcites{languageresource}{Language Resources}
\usepackage{graphicx}
\usepackage{tabularx}
\usepackage{soul}
\usepackage{booktabs}

\usepackage{epstopdf}
\usepackage[latin1]{inputenc}

\usepackage{hyperref}
\usepackage{xstring}

\title{\normalfont{\textbf{Advances in Pre-Training Distributed Word Representations}}}

\name{Tomas Mikolov, Edouard Grave, Piotr Bojanowski, Christian Puhrsch, Armand Joulin}

\address{Facebook AI Research \\
         \{tmikolov, egrave, bojanowski, cpuhrsch, ajoulin\}@fb.com\\}

\abstract{
Many Natural Language Processing applications nowadays rely on pre-trained word representations estimated
from large text corpora such as news collections, Wikipedia and Web Crawl. In this paper, we
show how to train high-quality word vector representations by using a combination of known tricks
that are however rarely used together. The main result of our work is the new set of publicly available pre-trained models
that outperform the current state of the art by a large margin on a number of tasks. \\ \newline }
\begin{document}

\maketitleabstract

\section{Introduction}

Pre-trained continuous word representations have become
basic building blocks of many Natural Language Processing (NLP) and
Machine Learning applications. 
These pre-trained representations provide distributional information 
about words,
that typically improve the generalization of models learned on limited amount of data~\cite{collobert2011natural}.
This information is typically derived from statistics
gathered from large unlabeled corpus of text data~\cite{deerwester1990indexing}.
A critical aspect of their training is thus to capture efficiently as much
statistical information as possible from rich and vast sources of data.

A standard approach for learning word representations is 
to train log-bilinear models based on either the 
skip-gram or the continuous bag-of-words (cbow) architectures, as
implemented in \emph{word2vec}~\cite{mikolov2013efficient} and \emph{fastText}~\cite{bojanowski2017enriching}\footnote{\url{https://fasttext.cc/}}.
In the skip-gram model, nearby words are predicted given a source word, while in the cbow model,
the source word is predicted according to its context.
These architectures and their implementation
have been optimized to produce high quality word 
representations able to transfer to many tasks,
while maintaining a sufficiently high training speed to scale to massive amount of data.


Recently, word2vec representations have been widely used 
in NLP pipelines to improve their performance.
Their impressive capability at transfering to new problems 
suggests that they are capturing important statistics about the 
training corpora~\cite{baroni2010distributional}.
As can be expected, the more data a model is trained on, the better the representations
are at transferring to other NLP problems.
Training such models on massive data sources, like Common Crawl, can be cumbersome and
many NLP practitioners prefer to use publicly available pre-trained word
vectors over training the models by themselves.
In this work, we provide new pre-trained word vectors that show consistent improvement
over the currently available ones, making them potentially very useful to a wide community
of researchers.


We show that several modifications of the standard word2vec training pipeline
significantly improves the quality of the resulting word vectors.
We focus mainly on known modifications and data 
pre-processing strategies that are rarely used together:
the position dependent features introduced by~\newcite{mnih2013learning},
the phrase representations used in~\newcite{mikolov2013distributed} and the use of subword information~\cite{bojanowski2017enriching}.

We measure their quality on standard benchmarks: syntactic, semantic and phrase-based
analogies~\cite{mikolov2013distributed},
rare words dataset~\cite{luong2013}, and as features in a question answering pipeline
on Squad question answering dataset~\cite{rajpurkar2016squad,chen2017reading}.

\section{Model Description}

In this section, we briefly describe the cbow model as it was used in word2vec, 
and then explain several known improvements to learn richer word representations.

\subsection{Standard cbow model}
\label{lab:cbow}

The cbow model as used in \newcite{mikolov2013efficient} learns word representations
by predicting a word according to its context. 
The context is defined as a symmetric window containing all the surrounding words.
More precisely, given a sequence of $T$ words $w_1,\dots,w_T$, the objective of 
the cbow model is to maximize the log-likelihood of the probability of 
the words given their surrounding, i.e.:
\begin{eqnarray}\label{eq:logp}
  \sum_{t=1}^T \log~p\left(w_t~|~C_t\right),
\end{eqnarray}
where $C_t$ is the context of the $t$-th word, e.g., the 
words $w_{t-c},\dots w_{t-1}, w_{t+1},\dots, w_{t+c}$ for a context window of size $2c$.
For now on, we assume that we have access to a scoring function between a word $w$ and its context $C$, denoted by $s(w, C)$. This scoring function
will be later parametrized by the word vectors, or representations.
A natural candidate for the conditional 
probability in Eq.~\ref{eq:logp} is a softmax function over the scores of a context
and words in the vocabulary. 
This choice is however impractical for large vocabulary.
An alternative is
to replace this probability by independent binary classifiers over words. 
The correct word is learned in contrast with a set of sampled negative candidates.
More precisely, the conditional probability of a word $w$ 
given its context $C$ in Eq.~(\ref{eq:logp}) is replaced by the following quantity:
\begin{eqnarray}\label{eq:prob}
  \log \left(1+ e^{-s(w,C)}\right) + \sum_{n\in N_C} \log\left(1+e^{s(n,C)}\right),
\end{eqnarray}
where $N_C$ is a set of negative examples sampled from the vocabulary. 
The objective function maximized by the cbow model is obtained by replacing the 
log probability in Eq.~(\ref{eq:logp}) by the quantity defined in Eq.~(\ref{eq:prob}), i.e.:
\begin{eqnarray*}\label{eq:cbow}
\sum_{t=1}^T \left[ \log \left(1+ e^{-s(w_t,C_t)}\right) + \sum_{n\in N_{C_t}} \log\left(1+e^{s(n,C_t)}\right)\right].
\end{eqnarray*}
A natural parametrization for this model is to represent each word $w$ by a vector $v_w$. 
Similarly, 
a context is represented by the average of word vectors $u_{w'}$ of each word $w'$ in its window.
The scoring function is simply the dot product between these two quantities, i.e.,
\begin{eqnarray}
s(w,C) = \frac{1}{|C|} \sum_{w'\in C} u_{w'}^Tv_w.
\end{eqnarray}
Note that different parametrizations are used for the words in a context and the predicted word.

\paragraph{Word subsampling.}
The word frequency distribution in a standard text corpus follows a Zipf distribution, which implies that most of the words belongs to small subset of the entire vocabulary~\cite{li1992random}.
Considering all the occurences of words equally would lead to overfit the parameters of the model on the representation of the most frequent words, while underfitting on the rest.
A common strategy introduced in \newcite{mikolov2013efficient} is to subsample frequent words, 
with the following probability $p_{disc}$ of discarding a word:
\begin{eqnarray}
  p_{disc}(w) = 1 - \sqrt{t / f_w} 
\end{eqnarray}
where $f_w$ is  the frequency of the word $w$, and $t>0$ is a parameter.


\subsection{Position-dependent Weighting}
\label{lab:pdw}

The context vector described above is simply the average of the word vectors contained in it. 
This representation is oblivious to the position of each word. 
Explicitly encoding a representation for both a word and its position would be impractical and prone to overfitting.
A simple yet effective solution introduced in the context of word representation by~\newcite{mnih2013learning} is to learn position representations
and use them to reweight the word vectors.
This position dependent weighting offers a richer context representation at a minimal 
computational cost.

Each position $p$ in a context window is associated with a vector $d_p$. The context vector
is then the average of the context words reweighted by their position vectors.
More precisely, denoting by $P$ the set of 
relative positions $[-c,\dots,-1,1,\dots,c]$ in the
context window, the context vector $v_{C}$ of the word $w_t$ is:
\begin{eqnarray}
  v_{C} = \sum_{p\in P} d_p \odot u_{t+p},
\end{eqnarray}
where $\odot$ is the pointwise multiplication of vectors.

\subsection{Phrase representations}

The original cbow model is only based on unigrams, which is 
insensitive to the word order.
We enrich this model with word n-grams to capture richer information.
Directly incorporating the n-grams in the models is quite challenging as
it clutters the models with uninformative content due to huge increase
of the number of the parameters.
Instead, we follow the approach of~\newcite{mikolov2013distributed}
where n-grams are selected by iteratively applying a mutual information
criterion to bigrams.
Then, in a data pre-processing step we merge the words in a selected n-gram into a single token.

For example, words with high mutual information like "New York"
are merged in a bigram token, "New\_York".
This  pre-processing step is repeated several times to form longer n-gram tokens,
like "New\_York\_City" or "New\_York\_University". 
In practice, we repeat this process $5 - 6$ times to build tokens representing longer ngrams.
We used the \emph{word2phrase} tool from the word2vec project\footnote{\url{https://github.com/tmikolov/word2vec}}.
Note that unigrams with high mutual information are merged
only with a probability of $50\%$, thus we still keep significant number of unigram occurrences.
Interstingly, even if the phrase representations are not further used in an application,
they effectively improve the quality of the word vectors, as is shown in the experimental section.



\subsection{Subword information}

Standard word vectors ignore word internal structure that contains
rich information. This information could be useful for computing representations
of rare or mispelled words, as well as for mophologically rich languages like Finnish or Turkish.
A simple yet effective approach is to enrich the word vectors with a bag of
character n-gram vectors that is either derived from the singular value
decomposition of the co-occurence matrix~\cite{schutze1993word} or directly
learned from a large corpus of data~\cite{bojanowski2017enriching}.
In the latter, each word is decomposed into its character n-grams $N$ and each n-gram $n$
is represented by a vector $x_n$. The word vector is then simply the sum of both
representations, i.e.:
\begin{eqnarray}
  v_w + \frac{1}{|N|} \sum_{n\in N} x_n.
\end{eqnarray}
In practice, the set of n-grams $N$ is restricted to the n-grams with $3$ to $6$ characters.
Storing all of these additional vectors is memory demanding.
We use the hashing trick to circumvent this issue~\cite{weinberger2009feature}.

\section{Training Data}

We used several sources of text data that are publicly available and the Gigaword
dataset, as described in Table 1. In particular, we used English Wikipedia from June 2017,
from which we used the meta pages archive which resulted in a text corpus with more than 9 billion words~\footnote{\url{https://dumps.wikimedia.org/enwiki/latest/}}.
Further, we used all news datasets from \url{statmt.org} from years 2007 - 2016, the UMBC corpus~\cite{umbc}, the English Gigaword,
and Common Crawl from May 2017\footnote{\url{https://commoncrawl.org/2017/06}}.

In case of the Common Crawl, we wrote a simple data extractor based on a unigram language model that
retrieves the documents written in English and discards low quality data. The same approach
can be in fact used to extract text data for many other languages from Common Crawl.

We decided to perform no complex data normalization or pre-processing, as we want the resulting
word vectors to be very easily used by a wide community (the text normalization can be done on top of the published word vectors
as a post-processing step). We only used a publicly available $tokenizer.perl$
script from the Moses MT project\footnote{\url{https://github.com/moses-smt}}. 
We observed that de-duplicating large text training corpora, especially Common Crawl,
significantly improves the quality of the resulting word vectors.

\begin{table}[!h]
\begin{center}
\begin{tabular}{ll}
  \toprule
 Corpus & Size [billion] \\
  \midrule
 Wikipedia meta-pages & 9.2 \\
 Statmt.org News & 4.2 \\
 UMBC News & 3.2 \\
 Gigaword & 3.3 \\
 Common Crawl & 630 \\
  \bottomrule
\end{tabular}
\caption{Training corpora and their size in billions of words after tokenization and sentence de-duplication.}
\end{center}
\end{table}


\section{Results}

Further we report results for models trained on either the Common Crawl, or on a combination of the Wikipedia, Statmt News, UMBC and Gigaword.
This is comparable to corpora that other models that attempted to improve upon word2vec were trained on, notably the GloVe model from
the Stanford NLP group~\cite{pennington2014glove}. Although a careful analysis performed in~\newcite{levy2015improving} shows that the original
word2vec is faster to train, produces more accurate models and takes significantly less memory than the GloVe algorithm,
the availability of large pre-trained GloVe models proved to be a useful resource for many researchers who do not have time
to train their own model on very large dataset like the Common Crawl.

We used the cbow architecture described in Section~\ref{lab:cbow} with window size 5 for the baseline models
and window size 15 for the models that learn position-dependent weights (described in Section~\ref{lab:pdw}).
We used 10 negative examples for training with the negative sampling
and threshold for subsampling frequent words set to $t = 10^{-5}$.

In Table~\ref{tab:crawl} we report results on the word analogies from~\newcite{mikolov2013efficient} using baseline cbow model trained on Common
Crawl with de-duplicated sentences, with phrases (we used 6 iterations of building the phrases by merging bigrams with high mutual information), and
with the position-dependent weighting as used in~\newcite{mnih2013learning}. The training itself
took three days on a single multi-core machine.

\begin{table}[!h]
\begin{center}
\begin{tabularx}{\columnwidth}{lXXX}
  \toprule
 Model & Sem & Syn & Tot \\
  \midrule
 cbow + uniq & 79 & 73 & 76 \\
 cbow + uniq + phrases & 82 & 78 & 80 \\
 cbow + uniq + phrases + weighting & 87 & 82 & 85 \\ 
  \bottomrule
 \end{tabularx}
\caption{Accuracies on semantic and syntactic analogy datasets for models trained on Common Crawl (630B words). By performing sentence-level de-duplication, adding position-dependent weighting and phrases, the model quality improves significantly.}
\label{tab:crawl}
\end{center}
\end{table}

In Table~\ref{tab:comparison}, we can see comparison between cbow as implemented in the fastText library~\cite{bojanowski2017enriching} and the GloVe
models trained on comparable corpora. The 87\% accuracy on the word analogy tasks is to our knowledge the best published result so far by a large margin,
and much better than existing GloVe models which were trained on comparable corpora. We improved this result further to 88.5\% accuracy by adding the sub-word features.

We also report very strong performance on the Rare Words dataset~\cite{luong2013},
again outperforming GloVe models by a large margin. Finally, we replaced the GloVe pre-trained vectors with the new fastText vectors in a
question answering system trained on the Squad dataset~\cite{rajpurkar2016squad}. In a setup that is further described in~\newcite{chen2017reading}, we did observe
significant improvement of the accuracy.

\begin{table}[!h]
\begin{center}
\begin{tabularx}{\columnwidth}{lXXX}
  \toprule
 Model & Analogy & RW & Squad \\
  \midrule
 GloVe Wiki + news & 72 & 0.38 & 77.7\% \\
 fastText Wiki + news & \bf{87} & 0.50 & 78.8\% \\
  \midrule
 GloVe Crawl & 75 & 0.52 & 78.9\% \\ 
 fastText Crawl & 85 & \bf{0.58} & \bf{79.8\%} \\
  \bottomrule
 \end{tabularx}
\caption{Results on Word Analogy, Rare Words and Squad datasets with fastText models trained on various corpora (see Table 1) or Common Crawl (see Table 2), and comparison to GloVe models trained on comparable datasets.}
\label{tab:comparison}
\end{center}
\end{table}

The models trained on the Wikipedia and News corpora, and on the Common Crawl, were published at the \url{fasttext.cc} website
and are available to the NLP researchers. Further, we did experiment with the phrase-based analogy dataset
introduced in~\newcite{mikolov2013distributed}, and achieved 88\% accuracy using the model trained on Crawl, which again
is to our knowledge the new state of the art result. We plan to release the model containing all the phrases in the near future.

\begin{table*}[t!]
  \centering
  \begin{tabular}{lccccccccccc}
    \toprule
    & Corpora & MRPC & MR & CR & SUBJ & MPQA & SST & TREC &~~& Average \\
    \midrule
    GloVe    & Wiki+news   & $71.9/81.0$  & $75.7$ & $78.1$ & $91.5$ & $86.9$ & $78.1$ & $66.6$ && $79.7$\\
    GloVe    & Crawl       & $72.0/80.7$  & $78.0$ & $79.6$ & $91.8$ & $88.0$ & $80.0$ & $84.2$ && $82.0$\\
    \midrule
    fastText & Wiki+news   & $72.9/\textbf{81.6}$  & $77.8$ & $80.3$ & $92.2$ & $\textbf{88.3}$ & $81.1$ & $\textbf{85.0}$ && $82.5$ \\
    fastText & Crawl       & $\textbf{73.4}/\textbf{81.6}$  & $\textbf{78.2}$ & $\textbf{81.1}$ & $\textbf{92.5}$ & $87.8$ & $\textbf{82.0}$ & $84.0$ && $\textbf{82.7}$\\
    \bottomrule
  \end{tabular}
  \caption{Comparison of different pre-trained models on supervised text classification tasks. }
  \label{tab:classif}
\end{table*}

Finally in Table~\ref{tab:classif}, we use a script provided by \newcite{conneau2017supervised} to
measure the influence of different pre-trained word vector models on several text classification tasks (MRPC, MR CR, SUBJ, MPQA, SST and TREC).
We performed the classification using the standard fastText toolkit running in a supervised mode~\cite{joulin2016bag}, using the pre-trained models
to initialize the classifier. Overall, the new fastText word vectors result in superior text classification performance.

\section{Discussion}

In this work, we have focused on providing very high quality set of pre-trained word and phrase vector representations.
Our findings indicate that improvements can be achieved by training well-known algorithms on very large text datasets, and that
using certain tricks can provide further gains in quality. Notably, we have found it very important to de-duplicate sentences
in large corpora such as the Common Crawl before training the models. Next, we have used an algorithm for building the phrases
in a pre-processing step. Finally, adding the position-dependent weights and subword features to the cbow model architecture gave us the final boost
of accuracy. The models described in this paper are freely available to researchers and engineers at the fastText webpage, and we hope
that these will be useful in various projects that use textual data.


\section{Acknowledgements}

We thank Marco Baroni and German Kruszewski for useful discussions and suggestions, Adam Fisch for help with the experiments on Squad
dataset, and Qun Luo for suggesting to use the position-dependent weighting at the word2vec discussion forum.

\section{Bibliographical References}
\label{main:ref}
\bibliographystyle{lrec}
\bibliography{e}


\end{document}